\documentclass[11pt]{article}
\usepackage{acl-hlt2011}
\usepackage{times}
\usepackage{latexsym}
\usepackage{graphicx}
\usepackage{amsmath}
\usepackage{amssymb}
\usepackage{mathrsfs}
\usepackage{multirow}
\usepackage{url}
\DeclareMathOperator*{\argmax}{arg\,max}
\title{A Comparative Study on Linguistic Feature Selection in Sentiment Polarity Classification}

\author{Zitao Liu \\
  University of Pittsburgh \\
  {\tt ztliu@cs.pitt.edu} \\}

\date{}

\begin{document}
\maketitle
\begin{abstract}
Sentiment polarity classification is perhaps the most widely studied topic. It classifies an opinionated document as expressing a positive or negative opinion. In this paper, using movie review dataset, we perform a comparative study with different single kind linguistic features and the combinations of these features. We find that the classic topic-based classifier(Naive Bayes and Support Vector Machine) do not perform as well on sentiment polarity classification. And we find that with some combination of different linguistic features, the classification accuracy can be boosted a lot. We give some reasonable explanations about these boosting outcomes.
\end{abstract}

\section{Introduction}

With the development of web technology, lots of user-generated content are available in the on-line documents. These text materials that come from many sources, like review site, discussion groups, contain people's opinions and attitudes, for example, whether a product review is positive or negative. Labeling those documents will not only enhance customers' satisfaction and experience, but also provide useful suggestions to those web companies. ~\cite{Pang:08} summarize that this classification task of labelling an opinionated document as expressing either an overall positive or an overall negative opinion is called sentiment polarity classification or polarity classification. Sentiment polarity classification would be helpful in information retrieval systems that can target accurate polarized documents based on opinioned queries. It would also be useful in business intelligence application and recommend system (e.g., ~\cite{Terveen:97,Tatemura:00}), where user input and feedback could be quickly summarized; indeed, in general, free-form survey responses given in natural language format could be processed using sentiment categorization. Moreover, there are also potential applications to message filtering; for example, one might be able to use sentiment information to recognize and discard ``flames" ~\cite{Spertus:97}.

In this paper, we review related work in Section~\ref{sec: related_work}, and then give several ways to choose features and illustrate the intuition and idea behind them (Section~\ref{sec: linguistic_features}). We choose two currently effective supervised learning classifier to make a prediction for classifying a review document as positive or negative (Section~\ref{sec: classifier}) and an evaluation corpus of 2000 documents is shown (Section~\ref{sec: evaluation}) and the results of basic linguistic features and some combinations of linguistic features are presented in Section~\ref{sec: evaluation}. Some conclusion and future work is presented in Section~\ref{sec: future}.

\section{Related Work}
\label{sec: related_work}

Much of research about automated sentiment and opinion detection has been performed these years. ~\cite{Wiebe:00} used the subjectivity of similar word as a confident weight when the classifier is not sure to automatically learn subjective adjectives from corpus, which provided good features for semantic application. ~\cite{Wiebe:01} described a straightforward method for automatically identifying collocational clues of subjectivity in texts and explored low-frequency words as features. ~\cite{Pang:02} tries several kinds of features in binary sentiment classification and made a comparative evaluation. ~\cite{Turney:03} extracted two consecutive words which contained at least one adjective or adverb and use PMI-IR to estimate the phrase's semantic orientation. ~\cite{Riloff:03} used AutoSlog-TS algorithm to learn subjective patterns based on two high-precision classifiers. ~\cite{Wilson:05} used both word features and polarity features for polarity classification. ~\cite{Cui:06} assumed that high order n-grams are more precise and deterministic expressions than unigrams or bigram, they employed high order n-grams to approximate surface patterns to capture the sentiment in text. ~\cite{Zhang:08} leveraged some document statistic features to improve the opinion polarity classification. ~\cite{Li:08} used character 3-grams as features based on assumptions that this can overcome spelling errors and problems of ill-formatted or ungrammatical questions and they made features by combine word and part-of-speech.

~\cite{Liu:10} work on polarity classification is perhaps the closest to ours. They explored some polarized features, transition features and some combinations with unigram. He applied maximum entropy classifier to evaluate those features by accuracy. In contrast, we utilize the contextual information of trigram and the effectiveness of adjective and adverb words, which we combine polarized features with both unigram and adjective/adverb trigram features. And we give an overall evaluation based on accuracy in different document representations and two classic classifiers.

\section{Linguistic Features}
\label{sec: linguistic_features}

We use supervised learning methods to achieve these sentiment polarity task: determine whether a review document is positive or negative. We use the subjective lexicon provide by ~\cite{Wilson:05}. It contains 2710 positive words and 4913 negative words. The features we explore are listed below.

\subsection{Ngram features}

The n-gram features we use are subsequences of n words from the original review documents. Here we choose n = 1, 2, 3 as features respectively as all of these three have been shown achieved good results in related work.

\subsection{Polarized unigram features}

Considering the word's polarity and part-of-speech are important for sentiment polarity classification. We combine those two information to make a new feature, represented as ``Polarity/Part-of-Speech". We collect our polarized unigram features in the following ways: for each unigram features in the corpus, we tag its polarity based on the subjective lexicon mentioned above (POS is for positive and NEG is for negative) and associate this polarity with its own part-of-speech. For example, ``I love that movie," the word ``love" is represent as ``POS/VB" which POS is for positive and VB is for verb. 

\subsection{Polarized bigram features}

Contextual information near the polarized words can be useful for sentiment polarity analysis, because the polarity of one word can be opposite given the word preceding itself. Considering those contextual information, we combine polarized unigram features with its preceding and following words' literal text or part-of-speech to represent this kind of contextual information. For example, ``I highly recommend this movie," since ``recommend" is a polarized unigram feature represented as ``POS/VB", we create four polarized bigram features: ``highly\_POS/VB", ``RB\_POS/VB", ``POS/VB\_this" and ``POS/VB\_DT".

\subsection{Adjective features}

Since people always express their emotion, subjectivity through adjective words and we assume that adjective words are more informative than others. Based on that intuition, we use those unigram words whose part-of-speech is JJ.

\subsection{Adjective/Adverb bigram features}

Considering that adverb words' occurrence also convey some private state and we also wants to capture some contextual information, we choose the a subset of features from bigram features: if there exists an adjective or adverb word in one bigram feature, we will keep that bigram as our adjective/adverb bigram feature; otherwise, discard it.

\subsection{Adjective/adverb trigram features}

Some researchers point out that the two consecutive words are not enough to capture that contextual information. For example, ``I recommend staying away from that movie" ``recommend" followed by ``staying away" implies the author's negative attitude towards this movie. So we explore those adjective/adverb trigram features in the following way: if there exists an adjective or adverb word in one trigram feature, we will keep that trigram as our adjective/adverb trigram feature; otherwise, discard it.

\subsection{Transition features}

Transition words, like ``however," ``contrarily" also lead to opposite sentiment polarity compared to the original sentence. This kind of information is not well explored in sentiment polarity analysis. In this study, we collect a transition list which contains 27 words or phrases. We construct the transition features in the following mechanism: if one sentence contains a transition word/phrase defined by the list, we take out all the noun, verb, adjective and adverb words in that sentence and their corresponding polarized unigram features. For example, in ``Although the director is famous \ldots " we use features like ``although\_director", ``although\_is", ``although\_famous", ``although\_POS/JJ" where ``POS/JJ" is the polarized unigram feature for word ``famous".

\section{Classifier}
\label{sec: classifier}

For this classification problem, we experiments our polarized documents with the two popular supervised learning algorithms: Naive Bayes and support vector machines. The optimization and learning principles are different in these two methods; however, both of them achieve effective results in traditional topic-based document classification.

To illustrate those two classification algorithm, let ${f_1,f_2,\ldots,f_m}$ be a predefined set of $m$ features that can appear in a document $doc$.

\subsection{Naive Bayes}

In naive bayes model, we determine a document's class based on the possibility that the document belongs to that class, namely given a document $doc$, its class label $c' = \argmax_c P(c|doc)$. And through the Bayesian rule, we can get
$$P(c|doc) = \frac{P(c)P(doc|c)}{P(doc)} \varpropto P(c)P(doc|c)$$
In Naive Bayes, we make the assumption that each features $f_i$ are conditionally independent given the class label. So we can get
$$P(c|doc) \varpropto P(c) \prod_{i=1}^m P(f_i|c) $$

Despite the assumption we made above is not held in the real word, the Naive Bayes model perform surprisingly well. ~\cite{Domingos:97} show that even with some dependent features, Naive Bayes is optimal for certain problems.

We used WEKA package(\url{http://www.cs.waikato.ac.nz/ml/weka/}) for training and testing, with all parameters set to their default values.  

\subsection{Support Vector Machine}

Support vector machines have been widely explored in classification area and it shows highly effective in traditional classification problems. Instead of applying probabilistic model like Naive Bayes, support vector machine tries to choose a hyperplane $\mathit{w}$ that separates the examples. Intuitively, a good separation is achieved by the hyperplane that has the largest distance to the nearest training data points of any class. In the binary classification scenario, we come to the optimization problem with constraints. Let $c_j \in \{ 1, -1 \}$ (corresponding to positive and negative) be the class label for document $doc_j$, the problem can be rewritten as
$$\mathit{w} = \sum_{j} \alpha_j \cdot c_j \cdot doc_j, \alpha_j \geq 0$$

We used ~\cite{Joachims:98} $SVM^{light}$ package for training and testing, with all parameters set to their default values.

\section{Evaluation}
\label{sec: evaluation}

\subsection{Data Collection}

In this study, we use the movie review data provided by Lillian Lee in \url{http://www.cs.cornell.edu/people/pabo/movie-review-data/}. It contains 1000 positive and 1000 negative processed movie reviews. Some statistics of this experimental data are shown in the Table~\ref{tab: corpus_statistics}.

\begin{table}[h]
\begin{center}
\begin{tabular}{|c|c|c|c|}
\hline \bf  & \bf Sentences & \bf Words & \bf Distinct  \\ 
 &  &  & \bf Words \\
\hline
Positive Docs & 31944 & 614970 & 35140 \\
\hline
Negative Docs & 33033 & 687664 & 37298\\
\hline
\end{tabular}
\end{center}
\caption{\label{tab: corpus_statistics} Statistic Information about Corpus. }
\end{table}

\begin{table*}[!hbpt]
\begin{center}
\begin{tabular}{|c|c|c|c|c|c|c|}
\hline 
\multirow{2}{*}{\bf No.} & \multirow{2}{*}{\bf Features} & \multirow{2}{*}{\bf \# of features} &  \multicolumn{2}{|c|}{\bf Naive Bayes} &  \multicolumn{2}{|c|}{\bf SVM}\\ 
\cline{4-7}
 & & & \bf Presence & \bf Frequency & \bf Presence & \bf Frequency\\ \hline 
 1 & Unigram(Non-Neg) & 1042107 & 0.807 &  0.677 & \bf 0.859 & 0.747\\\hline 
 2 & Unigram(Neg) & 1042107 & 0.795 &  0.684 & \bf 0.843 & 0.753\\\hline 
 3 & Bigram & 1040507 & 0.744 &  0.687 & \bf 0.815 & 0.781\\\hline 
 4 & Trigram & 1038907 & 0.704 &  0.672 & \bf 0.741 & 0.734\\\hline 
 5 & Adjective/Adverb & 503952 & 0.737 &  0.699 & 0.772 & \bf 0.773\\\hline 
 6 & Adjective & 104164 & 0.774 &  0.716 & \bf 0.802 & 0.792\\\hline 
 7 & Polarized Bigram & 625734 & 0.776 &  0.708 & \bf 0.794 & 0.783\\\hline 
 8 & Polarized Unigram & 156671 & 0.558 & 0.628 & 0.560 & \bf 0.735\\\hline 
 9 & Adjective/Adverb Trigram & 438120 & 0.653 &  0.639 & \bf 0.702 & \bf 0.702\\\hline 
\end{tabular}
\end{center}
\caption{\label{tab: basic_feature} Average five-fold cross-validation accuracies. Boldface: best performance for a given setting(row). Neg means apply negation tagging and Non-Neg means without negation tagging}
\end{table*}

\subsection{Experimental Setting}

Before we applying the features discussed in Section~\ref{sec: linguistic_features}, we extended all the negation abbreviation to its original form, like changing ``It isn't$\ldots$" to ``It is not $\ldots$" Then we removed all the punctuations from the corpus except the exclamation mark and question mark. We adapt a technique of ~\cite{Das:01}, we added the tag \textbf{NOT\_} to every word between the negation word ``not". We use OpenNLP POS tagger(\url{http://opennlp.sourceforge.net}) to tag words' part-of-speech. We use both the binary model(presence) and vector space model(frequency) to represent each document respectively. In binary model, each dimension of the document vector is 0 or 1 based on whether this feature appears or not. In vector space model, each dimension of the document vector is the number of occurrence of that feature in this document.  Also we set the term-removal threshold to 5, which means that we remove all the features   whose frequency is lower than 5. We apply 5-cross validation for each feature selection methods.

\subsection{Basic Linguistic Features Result   }

According to Table ~\ref{tab: basic_feature} we can see that the highest accuracy we can achieve is 0.859 with boolean model representation using SVM. However, both these two classifiers have been reported to achieve more than 0.9 accuracy using unigram features ~\cite{Joachims:98} in traditional topic-based classification. This also  indicates that sentiment polarity classification is more difficult than topic-based classification and researchers should put more attention on this area.

On the other hand, we can see than SVM classifier has a better performance than our baseline classifier Naive Bayes for all different features, which is also an evidence that the way of trying to find the maximum margin hyperplane is more robust than the simple independence assumption in Naive Bayes.

In both Naive Bayes and SVM classifier, unigram features achieve the best accuracy in presence representation and the Adjective features is the highest one in frequency representation. The reason is in presence representation, each feature is only counted as 1 or 0, which means treating polarized features and the ordinary features as the same weight. This may cause some loss of classification information. And unigram has the most number of features and highest probability frequency for each feature, which is resistant to the information loss in presence representation model. In frequency representation model, people always express their emotion, sentiment through adjectives, the adjective feature's counts play an important role in features. The term frequency is more beneficial for those adjective features which make the adjective features have the highest accuracy in term frequency representation.

\textbf{Frequency vs. Presence   }In Table ~\ref{tab: basic_feature}, the results from Naive Bayes classifiers, except the polarized unigram features, all the other features have better accuracies in presence representation than frequency representation. And these presence's better accuracies also appear in SVM classifiers. Interestingly, this is a direct opposition to the observations of ~\cite{McCallum:98}. And our experimental results also verify ~\cite{Pang:02}'s speculation that the difference of accuracies in presence and frequency is an indication the difference between sentiment polarity classification and topic-based classification.

\subsection{Combined Linguistic Features Results   }

Instead of simply evaluate some basic linguistic features in Table ~\ref{tab: basic_feature}, we want to explore some combination of different features, which may lead a better classification result.

\textbf{Combined Unigram Features   }
Since the unigram features only treat each word as a features, this may lose some contextual information in classification process, we try to combine the Polarized Unigram features(PU), Polarized Bigram features(PB) and Transition Features(T) respectively. In totally, we have seven combination with the original unigram features.

\begin{figure}[t]
\begin{center}
\includegraphics[width=0.5\textwidth]{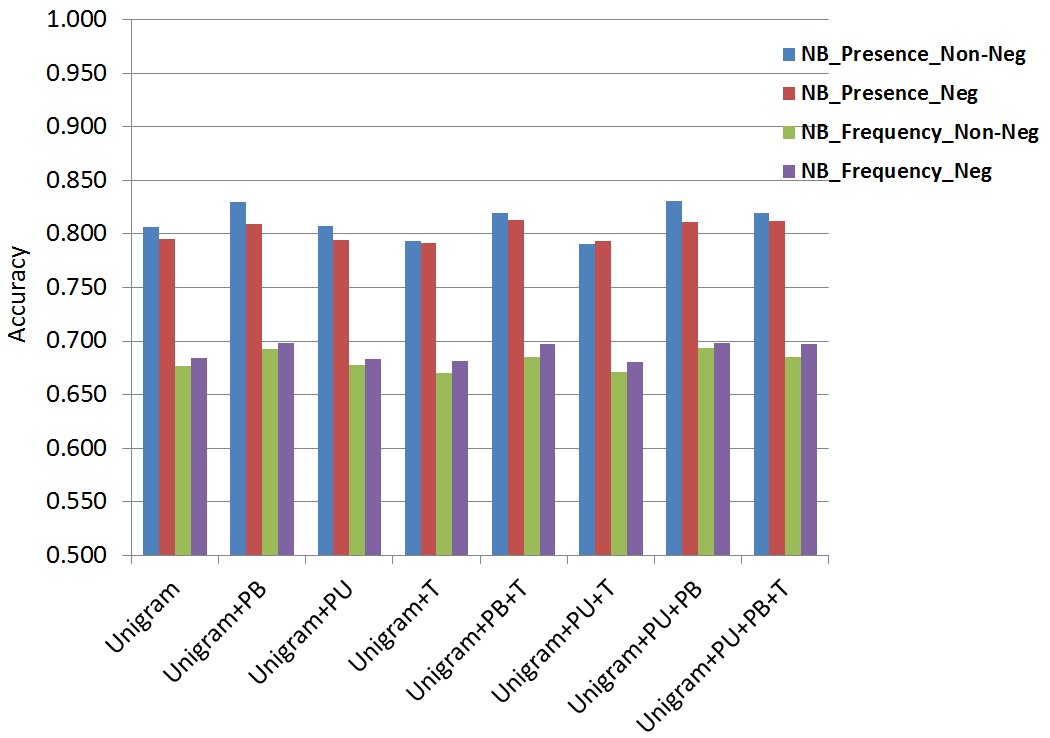}
\caption{Unigram combination features's accuracy in Naive Bayes classifier. \emph{NB\_Presence\_Non-Neg} means the Naive Bayes' classification accuracy for unigram features without applying negation rules in presence representation.}
\label{fig: nb_neg_or_non}
\end{center}
\end{figure}

\begin{figure}[t]
\begin{center}
\includegraphics[width=0.5\textwidth]{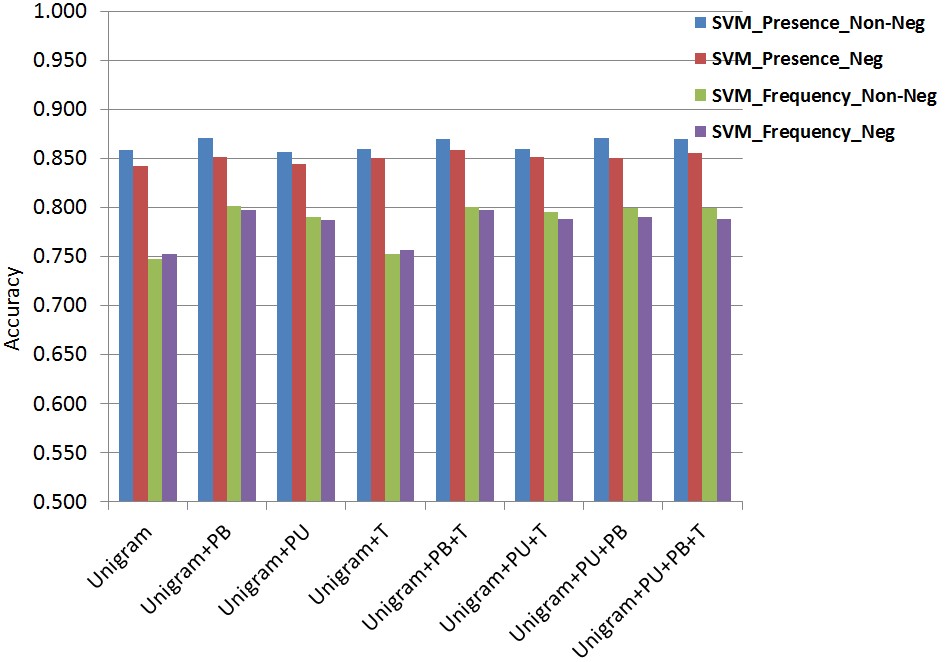}
\caption{Unigram combination features's accuracy in SVM classifier. \emph{SVM\_Presence\_Non-Neg} means the SVM' classification accuracy for unigram features without applying negation rules in presence representation.}
\label{fig: svm_neg_or_non}
\end{center}
\end{figure}

\textbf{Negation vs. Non-negation   }
We try two versions of original unigrams: unigram with negation and unigram without negation. See the accuracies of these two kind of versions' features by using Naive Bayes classifier in Figure ~\ref{fig: nb_neg_or_non} and by using SVM classifier in Figure ~\ref{fig: svm_neg_or_non}. In presence representation, we can see that both Naive Bayes and SVM classifier achieve a higher accuracy without applying the negation rules reported in ~\cite{Das:01}. However, the observations are changed in the term frequency representation. Negation version features beats the non-negation features in Naive Bayes features in all the different combinations while in SVM classifier, only unigram features and Unigram+T features has a higher accuracy. Interestingly, this is in direct opposition to the observations of ~\cite{Pang:02}.

\begin{figure}[t]
\begin{center}
\includegraphics[width=0.5\textwidth]{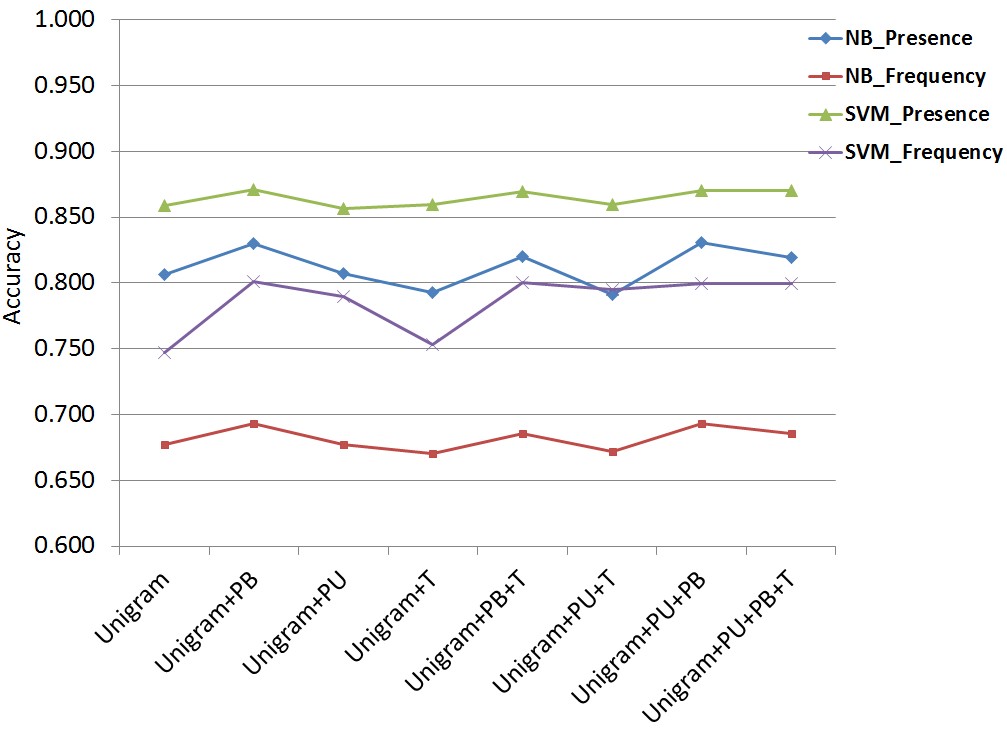}
\caption{Non-negation unigram combination features's accuracy in Naive Bayes and SVM classifier}
\label{fig: unigram}
\end{center}
\end{figure}

Based on the above observation, we only analyse the combination with non-negation unigram features. From the Figure ~\ref{fig: unigram}, we can see that the presence representation nearly achieves all the better accuracies than the frequency representation in different combination with unigram. This is also a verification of our analysis in the previous section. Also we can see that Unigram+PU+PB achieves the highest accuracy in \emph{NB\_Presence}, Unigram+PB, Unigram+PU+PB achieves the highest accuracy in \emph{NB\_Frequency} and \emph{SVM\_Presence} and Unigram+PB, Unigram+PB+T achieves the highest accuracy in \emph{SVM\_Frequency}. All of these above features contain the Polarized Bigram(PB) part. We think this is (1)unigram inevitably loses the contextual information; (2)polarized bigram features capture the polarized contextual information, which is more important to sentiment polarity classification. Even these polarized contextual information is a small subset of entire contextual information, but these polarized information is the most decisive part.

\textbf{Combined Adjective/Adverb Trigram Features   }
In Table ~\ref{tab: basic_feature}, we can see that Adjective/Adverb Trigram features nearly has the second lowest accuracy no matter which representation model and classifier are used. We think this because even the adjective/adverb trigram features capture enough contextual information for classifier, but each of these features' appearances in each document are very low, which make the document representation vector very sparse. This leads to the low accuracy. So we combine some short polarized features to boost the accuracy of adjective/adverb trigram features. We use 3AdjAdv to denote adjective/adverb trigram features for short.

\begin{figure}[t]
\begin{center}
\includegraphics[width=0.5\textwidth]{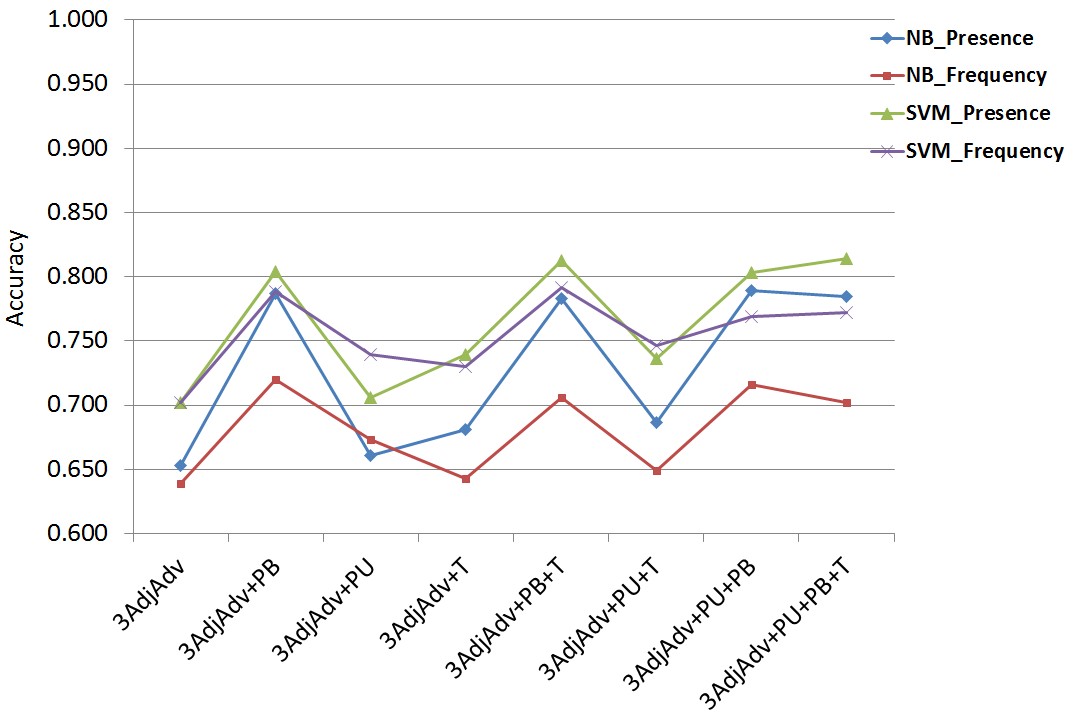}
\caption{Adjective/Adverb trigram combination features's accuracy in Naive Bayes and SVM classifier}
\label{fig: 3AdjAdv}
\end{center}
\end{figure}

From the Figure ~\ref{fig: 3AdjAdv}, we can see that (1) when adding Polarized Unigram features(PU) or Transition features(T) separately, the accuracy is improved slightly. (2) Polarized Bigram features(PB) boosts the accuracy hugely compared to PU and T. 3AdjAdv+PB features achieves the highest accuracies in different representations and classifiers. (3) The combination of PB and T features are more informative to SVM classifier, which achieves the highest accuracies in both presence and frequency representation. We think this is because these three consecutive words features are much more sensitive to the sentiment transition in the sentence. Without the transition information, adjective/adverb trigram features around the transition words are self-contradictory and is harmful for the accuracy of classification. 

\section{Conclusions}
\label{sec: future}

In this paper, we make a comparative study of effectiveness of different kinds of linguistic features. We give an overview evaluation of these linguistic features based on sentiment polarity classification accuracy. We also explore the question: whether the combination of basic linguistic features can boost the accuracy of sentiment polarity classification. We try eight combinations on two different kinds of basic linguistic features in different document representation and classifiers and we find that Polarized Bigram features(PB) contains enough polarized contextual information, which is the most informative features for accuracy boosting. And we also find the for trigram features, Transition features(T) are also very important in accuracy improvement since these features correct the inaccurate features in trigrams that around the sentiment transition.

\end{document}